%% file: TPAMI-GPaCo.tex
\documentclass[10pt,journal,compsoc]{IEEEtran}
\usepackage{pifont}

\usepackage{ifpdf}

\ifCLASSOPTIONcompsoc
\usepackage[nocompress]{cite}
\else
\usepackage{cite}
\fi
\ifCLASSINFOpdf
\usepackage[pdftex]{graphicx}
\else
\usepackage[dvips]{graphicx}
\fi
\usepackage{amsmath}
\usepackage{algorithmic}

\usepackage{array}

\ifCLASSOPTIONcompsoc
\usepackage[caption=false,font=footnotesize,labelfont=sf,textfont=sf]{subfig}
\else
\usepackage[caption=false,font=footnotesize]{subfig}
\fi
\usepackage{url}

\usepackage{color}
\usepackage{lipsum}
\usepackage{algorithm}
\newtheorem{theorem}{Remark}
\usepackage{multicol}
\usepackage{multirow}
\usepackage{booktabs}
\usepackage[accsupp]{axessibility}
\usepackage[marginal]{footmisc}
\newlength\savewidth

\usepackage{ragged2e}

\begin{document}

	\title{Generalized Parametric Contrastive Learning}
	\author{Jiequan~Cui,~\IEEEmembership{Student Member,~IEEE,}
	    Zhisheng~Zhong,~\IEEEmembership{Student Member,~IEEE,}
	    Zhuotao~Tian,~\IEEEmembership{Student Member,~IEEE,}
		Shu~Liu,~\IEEEmembership{Member,~IEEE,}
		Bei~Yu,~\IEEEmembership{Member,~IEEE,} 
		Jiaya~Jia,~\IEEEmembership{Fellow,~IEEE} 
		
		\IEEEcompsocitemizethanks{\IEEEcompsocthanksitem J.~Cui, Z.~Zhong, Z.~Tian, B.~Yu and J.~Jia are with the Department of Computer Science \& Engineering, The Chinese University of Hong Kong, ShaTin, Hong Kong.\protect\\
			E-mail: \{jiequancui, tianzhuotao, liushuhust\}@gmail.com \\ \{zszhong21, byu, leojia\}@cse.cuhk.edu.hk;  
			\IEEEcompsocthanksitem J.~Jia and S.~Liu are with the SmartMore.}}

	\IEEEtitleabstractindextext{%
		\begin{abstract}
            \justifying In this paper, we propose the Generalized Parametric Contrastive Learning (GPaCo/PaCo) which works well on both imbalanced and balanced data. Based on theoretical analysis, we observe supervised contrastive loss tends to bias on high-frequency classes and thus increases the difficulty of imbalanced learning. We introduce a set of parametric class-wise learnable centers to rebalance from an optimization perspective. Further, we analyze our GPaCo/PaCo loss under a balanced setting. Our analysis demonstrates that GPaCo/PaCo can adaptively enhance the intensity of pushing samples of the same class close as more samples are pulled together with their corresponding centers and benefit hard example learning. 
			Experiments on long-tailed benchmarks manifest the new state-of-the-art for long-tailed recognition. 
			On full ImageNet, models from CNNs to vision transformers trained with GPaCo loss show better generalization performance and stronger robustness compared with MAE models.
			Moreover, GPaCo can be applied to semantic segmentation task and obvious improvements are observed on 4 most popular benchmarks.
			Our code is available at {\bf \url{ https://github.com/dvlab-research/Parametric-Contrastive-Learning}}.
		\end{abstract}
		\begin{IEEEkeywords}
			Representation Learning, Contrastive Learning, OOD Robustness, Long-tailed Recognition, Semantic Segmentation.
	\end{IEEEkeywords}}

    \maketitle
    \IEEEdisplaynontitleabstractindextext
    \IEEEpeerreviewmaketitle

\input{doc/intro}

\input{doc/related}

\input{doc/algo}

\input{doc/exp}

\input{doc/conclu}

\ifCLASSOPTIONcaptionsoff
\newpage
\fi

{\small
\bibliographystyle{IEEEtran}
\bibliography{egbib}
}

\end{document}

%% file: doc/intro.tex
\IEEEraisesectionheading{\section{Introduction}\label{sec:introduction}}

\IEEEPARstart{C}{onvolutional} neural networks (CNNs) have achieved great success in various tasks, including image classification \cite{he2016deep,vggnet}, object detection \cite{DBLP:conf/cvpr/LinDGHHB17, DBLP:conf/cvpr/LiuQQSJ18} and semantic segmentation \cite{DBLP:conf/cvpr/ZhaoSQWJ17}. Especially, with the rise of neural network search \cite{DBLP:conf/cvpr/ZophVSL18, DBLP:conf/iclr/LiuSY19, DBLP:conf/cvpr/TanCPVSHL19, DBLP:conf/iccv/CuiCLLSJ19, DBLP:conf/iclr/CaiGWZH20}, performance of CNNs have further taken a big step. However, The huge progress highly depends on large-scale and high-quality datasets, such as ImageNet \cite{imagenet}, MS COCO \cite{coco} and Places \cite{zhou2017places}. But when deal with real-world applications, generally we face the long-tailed distribution problem -- a few classes contain many instances, while most classes contain only a few instances. Learning in such an imbalanced setting is challenging as the low-frequency classes can be easily overwhelmed by high-frequency ones. Without considering this situation, CNNs will suffer from significant performance degradation. 

\begin{figure*}[ht]
\centering
\subfloat[Comparison on ImageNet-LT for long-tailed image classification.]{
    \includegraphics[width=0.48\textwidth]{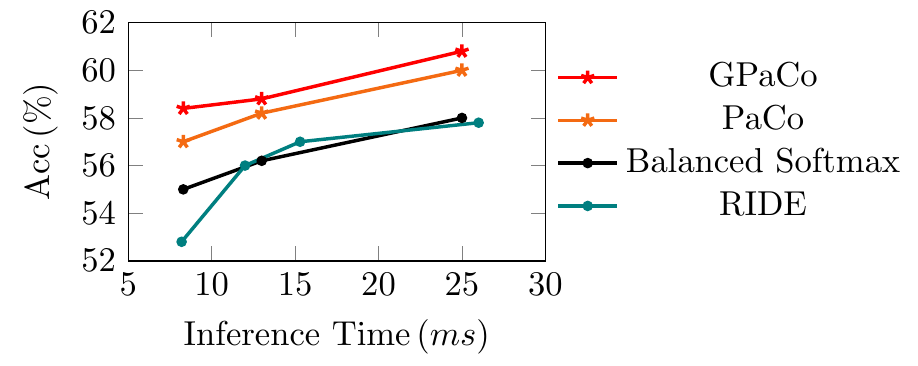}%
}
\hspace{0.05in}
\subfloat[Comparison on ADE20K for semantic segmentation.]{
    \vspace{-0.2in}
    \includegraphics[width=0.48\textwidth]{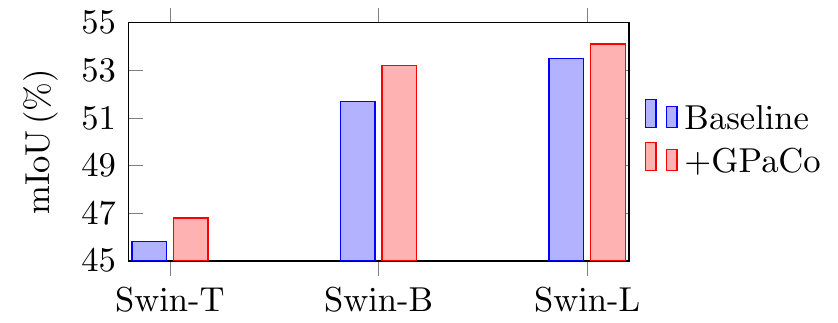} %
}
\vspace{0.05in}
\subfloat[Comparison with MAE \cite{he2022masked} ViT models on full ImageNet and out-of-distribution robustness.]{
    \begin{tabular}{lccccc}
    \toprule
        Models & ImageNet($\uparrow$) & ImageNet-C(mCE $\downarrow$) &ImageNet-C(rel. mCE $\downarrow$) &ImageNet-R($\uparrow$) &ImageNet-S($\uparrow$) \\
        \midrule
        ViT-B &+0.4 &-1.9 &-2.6 &+1.8 &+3.3\\
        \midrule
        ViT-L &+0.3 &-1.7 &-2.4 &+0.0 &+2.8\\
        \bottomrule
        \end{tabular}
}
\caption{\textbf{GPaCo demonstrates impressive performance on imbalanced and balanced data.} (a) shows that GPaCo achieves state-of-the-art on long-tailed classification. GPaCo benefits semantic segmentation task and significant improvement is observed on ADE20K in (b). In (c), top-1 accuracy are reported for full ImageNet, ImageNet-R, and ImageNet-S. mCE and rel. mCE are used for ImageNet-C. Compared with MAE models, GPaCo enjoys better generalization ability and stronger robustness on ImageNet and its variants.}
\label{fig:comparison}
\end{figure*}

\begin{figure}[t]
	\begin{center}
		\includegraphics[width=0.98\linewidth]{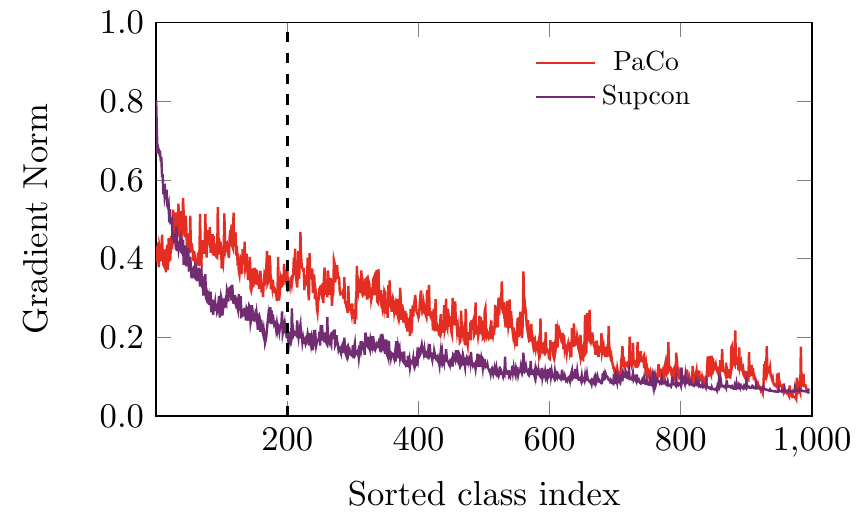}
		\caption{\textbf{Rebalance in contrastive learning.} We collect the average L2 norm of the gradient of weights in the last classifier layer on ImageNet-LT. Category indices are sorted by their image counts. The gradient norm varying from the most frequent class to the least one is steep for supervised contrastive learning \cite{DBLP:conf/nips/KhoslaTWSTIMLK20}. In particular, the gradient norm dramatically decreases for the top 200 most frequent classes. Trained with PaCo, the gradient norm is better balanced.}
		\label{fig:gradient}
	\end{center}
	\vspace{-0.2in}
\end{figure}

Contrastive learning \cite{DBLP:conf/icml/ChenK0H20, DBLP:conf/cvpr/He0WXG20, DBLP:journals/corr/abs-2003-04297, DBLP:conf/nips/GrillSATRBDPGAP20, DBLP:conf/nips/CaronMMGBJ20} is a major research topic due to its success in self-supervised representation learning. Khosla~\cite{DBLP:conf/nips/KhoslaTWSTIMLK20} extend non-parametric contrastive loss into non-parametric supervised contrastive loss by leveraging label information, which trains representation in the first stage and learns the linear classifier with the fixed backbone in the second stage. Though supervised contrastive learning works well in a balanced setting, for imbalanced datasets, our theoretical analysis shows that high-frequency classes will have a higher lower bound of loss and contribute much higher importance than low-frequency classes when equipping it in training.
This phenomenon leads to model bias on high-frequency classes and increases the difficulty of imbalanced learning.
As shown in Fig.~\ref{fig:gradient}, when the model is trained with supervised contrastive loss on ImageNet-LT, the gradient norm varying from the most frequent class to the least one is rather steep. In particular, the gradient norm dramatically decreases for the top 200 most frequent classes.    

Previous work \cite{DBLP:journals/nn/BudaMM18,DBLP:conf/cvpr/HuangLLT16,DBLP:conf/cvpr/CuiJLSB19, he2009learning,chawla2002smote, shen2016relay, DBLP:conf/nips/RenYSMZYL20, DBLP:conf/iclr/KangXRYGFK20, DBLP:journals/corr/abs-2010-01809, DBLP:journals/corr/abs-2101-10633, DBLP:conf/nips/TangHZ20, DBLP:journals/corr/abs-2101-10633, Zhong_2021_CVPR, cui2022region} explored to rebalance in traditional supervised cross-entropy learning. In this paper, we tackle the above mentioned imbalance issue in supervised contrastive learning and make use of contrastive learning for long-tailed recognition. 

To rebalance in supervised contrastive learning, we introduce a set of parametric class-wise learnable centers into supervised contrastive learning. We name our algorithm \textbf{Parametric Contrastive Learning (PaCo)} shown in Fig.~\ref{fig:paco_psum} (a). With such a simple and yet effective operation, we theoretically prove that the optimal values for the probability that two samples are a true positive pair (belonging to the same class), varying from the most frequent class to the least frequent class, are more balanced. Thus their lower bound of loss values are better organized. This phenomenon means the model takes more care of low-frequency classes, making the PaCo loss benefit imbalanced learning. Fig.~\ref{fig:gradient} shows that, with our PaCo loss in training, gradient norm varying from the most frequent class to the least one are moderated better than supervised contrastive learning, which matches our analysis. 

Further, we analyze the PaCo loss under a balanced setting. Our analysis demonstrates that with more samples clustered around their corresponding centers in training, the PaCo loss increases the intensity of pushing samples of the same class close, which benefits hard examples learning. 

MoCo \cite{DBLP:conf/cvpr/He0WXG20} enables small mini-batch training for self-supervised contrastive learning with a momentum encoder and a queue. Chen et al. \cite{chen2021exploring} claims that it is the ``stop-gradient" but not the momentum encoder that is necessary to avoid collapsing solutions for self-supervised learning. We examine the effects of each component in PaCo including augmentation strategy, loss function (relation to multi-task), momentum encoder and queue size. Interestingly, the simplified \textbf{Generalized Parametric Contrastive Learning (GPaCo)} by removing the momentum encoder even boosts model performance and robustness, indicating the unnecessary of momentum encoder in PaCo framework.

Finally, we conduct experiments on imbalanced data including long-tailed version of CIFAR \cite{cb-focal, DBLP:conf/nips/CaoWGAM19}, ImageNet \cite{Liu_2019_CVPR}, Places \cite{Liu_2019_CVPR} and iNaturalist 2018 \cite{van2018inaturalist}. Experimental results show that we create a new record for long-tailed recognition. 
On balanced data, We testify the effectiveness of GPaCo on full ImageNet \cite{imagenet} and CIFAR \cite{krizhevsky2009learning}. Compared with MAE models, we achieve better generalization ability and stronger robustness. With GPaCo loss for semantic segmentation, obvious improvements are obtained on popular benchmarks including ADE20K \cite{zhou2017scene}, COCO-Stuff \cite{caesar2018coco}, PASCAL Context \cite{mottaghi2014role} and Cityscapes \cite{cordts2016cityscapes}. GPaco demonstrates its great generality with various tasks as shown in Fig~\ref{fig:comparison}.
Our key contributions are as follows.
\begin{itemize}
	\item We identify the shortcoming of supervised contrastive learning under an imbalanced setting -- it tends to bias high-frequency classes.
	\item We extend supervised contrastive loss to the PaCo loss, which is more friendly to imbalanced learning, by introducing a set of parametric class-wise learnable centers.
	\item We examine the necessary of components in PaCo and simplify PaCo to GPaCo, observing that the momentum encoder can hurt model performance and should be removed.
	\item GPaCo can benefit the training on imbalanced data and balanced data varying from CNNs to vision transformers. Experiments on tasks {\it e.g.}, long-tailed classification, full ImageNet/CIFAR classifcation, out-of-distribution robustness, and semantic segmentation demonstrate the generality of GPaCo.
\end{itemize}

%% file: doc/related.tex
\section{Related Work}

\noindent{\bf Re-sampling/re-weighting.}
The most classical way to deal with long-tailed datasets is to over-sample low-frequency class images \cite{shen2016relay, zhong2016towards, buda2018systematic, byrd2019effect}
or under-sample high-frequency class images \cite{he2009learning, japkowicz2002class,buda2018systematic}. However, Oversampling can suffer from heavy over-fitting to low-frequency
classes especially on small datasets. For under-sampling, discarding a large portion of high-frequency class data inevitably causes degradation of the generalization ability of CNNs. Re-weighting \cite{huang2016learning, huang2019deep,wang2017learning,ren2018learning,shu2019meta,jamal2020rethinking} the loss functions is an alternative way to rebalance by either enlarging weights on more challenging and sparse classes or randomly ignoring gradients from high-frequency classes \cite{tan2020eql}. However, with large-scale data, re-weighting makes CNNs difficult to optimize during training \cite{huang2016learning, huang2019deep}.

\vspace{+0.1in}
\noindent{\bf One/two-stage Methods.}
Since deferred re-weighting and re-sampling were proposed by Cao ~\cite{DBLP:conf/nips/CaoWGAM19}, Kang~\cite{DBLP:conf/iclr/KangXRYGFK20} and Zhou~\cite{zhou2019bbn} observed re-weighting or re-sampling strategies could benefit classifier learning while hurting representation learning. Kang~\cite{DBLP:conf/iclr/KangXRYGFK20} proposed to decompose representation and classifier learning. It first trains the CNNs with uniform sampling, and then fine-tune the classifier with class-balanced sampling while keeping parameters of representation learning fixed. Zhou~\cite{zhou2019bbn} proposed one cumulative learning strategy, with which they bridge representation learning and classifier re-balancing.

The two-stage design is not for end-to-end frameworks.
Tang~\cite{DBLP:conf/nips/TangHZ20} analyzed the reason from the perspective of causal graph and concluded that the bad momentum causal effects played a vital role.
Cui ~\cite{DBLP:journals/corr/abs-2101-10633} proposed residual learning mechanism to address this issue.

\vspace{+0.1in}
\noindent{\bf Non-parametric Contrastive Loss.}
\label{Sec:contrastive_learning}
Contrastive learning \cite{DBLP:conf/icml/ChenK0H20, DBLP:conf/cvpr/He0WXG20, DBLP:journals/corr/abs-2003-04297, DBLP:conf/nips/GrillSATRBDPGAP20, DBLP:conf/nips/CaronMMGBJ20} is a framework that learns similar/dissimilar representations from data that are organized into similar/dissimilar pairs. An effective contrastive loss function, called InfoNCE \cite{DBLP:journals/corr/abs-1807-03748}, is
\begin{equation}
\mathcal{L}_{q, k^+, \{k^-\}} = -\log \frac{\exp(q{\cdot}k^+ / \tau)}{\exp(q{\cdot}k^+ / \tau) + {\displaystyle\sum_{k^-}}\exp(q{\cdot}k^-  / \tau)},
\label{eq:infonce}
\end{equation}
where $q$ is a query representation, $k^+$ is for the positive (similar) key sample, and $\{k^-\}$ denotes negative (dissimilar) key samples. $\tau$ is a temperature hyper-parameter. In the {instance discrimination} pretext task \cite{DBLP:conf/cvpr/WuXYL18} for self-supervised learning, a query and a key form a positive pair if they are data-augmented versions of the same image. It forms a negative pair otherwise. 

Traditional cross-entropy with linear fc layer weight $w$ and true label $y$ among $n$ classes is expressed as 
\begin{equation}
\mathcal{L}_{cross-entropy} = -\log \frac{\exp(q{\cdot} w_{y})}{\sum_{i=1}^{n} \exp(q{\cdot}w_{i})}.
\label{eq:cross-entropy}
\end{equation}

Compared to it, InfoNCE does not get involved with parametric learnable parameters.
To distinguish our proposed parametric contrastive learning from previous ones, we treat the InfoNCE as a non-parametric contrastive loss following \cite{wu2018unsupervised}.

Chen~\cite{DBLP:conf/icml/ChenK0H20} used self-supervised contrastive learning SimCLR to first match the performance of a supervised ResNet-50 with only a linear classifier trained on self-supervised representation on full ImageNet.
He~\cite{DBLP:conf/cvpr/He0WXG20} proposed MoCo and Chen ~\cite{DBLP:journals/corr/abs-2003-04297} extended MoCo to MoCo v2, with which small batch size training can also achieve competitive results on full ImageNet \cite{imagenet}.
In addition, many other methods \cite{DBLP:conf/nips/GrillSATRBDPGAP20, DBLP:conf/nips/CaronMMGBJ20} are also proposed to further boost performance.

%% file: doc/algo.tex
\section{Parametric Contrastive Learning}

\begin{figure*}[t]
		\centering
			\subfloat[Parametric contrastive learning (PaCo).]{
			\includegraphics[width=.616\textwidth]{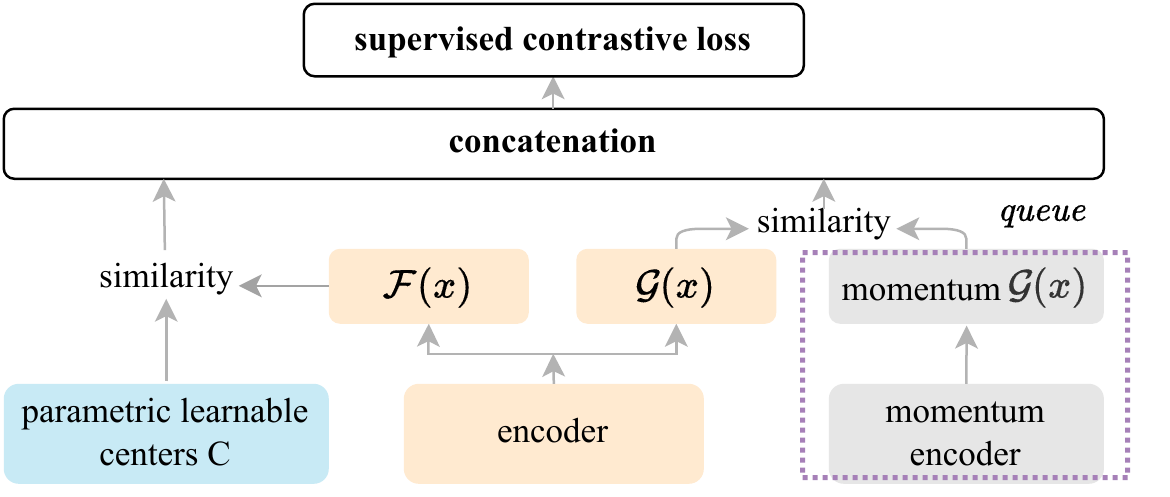}%
		}
	\vspace{.1in}
	\subfloat[Curve for $\mathcal{L}_{extra}$.]{
			\includegraphics[width=.32\textwidth]{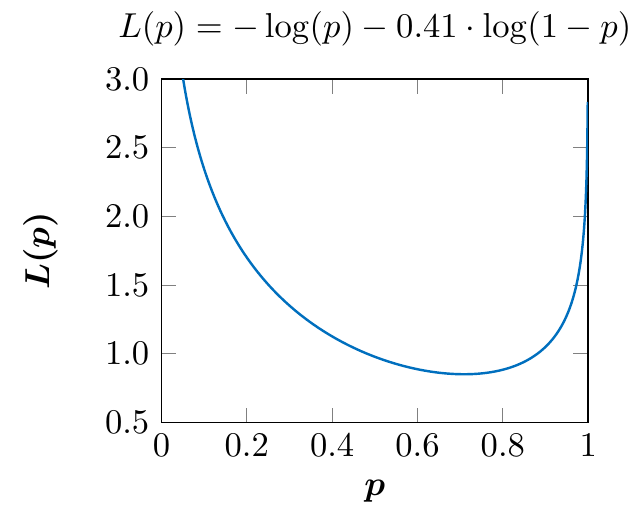}%
		}
		\caption{\textbf{Framework of Generalized Parametric Contrastive Learning (GPaCo/PaCo) and $L_{extra}$ curve.} (a) illustrates the details of GPaCo/PaCo. PaCo incorporates a momentum encoder, whereas GPaCo eliminates it. We introduce a set of parametric class-wise learnable centers for rebalancing in contrastive learning. More analysis is in Sec.~\ref{Sec:PaCo} for PaCo and Sec.~\ref{sec:gpaco} for GPaCo. (b) plots the curve of $L_{extra}$ instantiated with full ImageNet. More analysis on balanced data is in Section \ref{sec:PaCo_balance}.}
		\label{fig:paco_psum}

\end{figure*}

\subsection{Supervised Contrastive Learning}
\label{Sec:supcon}

Khosla~\cite{DBLP:conf/nips/KhoslaTWSTIMLK20} extended the self-supervised contrastive loss with label information into supervised contrastive loss. Here we present it in the framework of MoCo \cite{DBLP:conf/cvpr/He0WXG20, DBLP:journals/corr/abs-2003-04297} as
\begin{equation}
\mathcal{L}_{i} \!=\! -\sum_{ z_{+} \in P(i)} \log \frac{ \exp({z_{+} \cdot T(x_{i})}) }{\sum_{z_{k} \in A(i)} \exp({z_{k} \cdot T(x_{i})})}. \label{eq:supcon} \\
\end{equation}
MoCo framework \cite{DBLP:conf/cvpr/He0WXG20, DBLP:journals/corr/abs-2003-04297} consists of two networks with the same structure, {\it i.e.}, { \it query network} and {\it key network}. The key network is driven by a momentum update with the query network in training. For each network, it usually contains one encoder CNN and one two-layer MLP transform.

During training, for one two-viewed image batch $B=(B_{v1}, B_{v2})$ and label $y$, $B_{v1}$ and $B_{v2}$ are fed into the {query network} and {key network} respectively and we denote their outputs as $Z_{v1}$ and $Z_{v2}$.
Especially, $Z_{v2}$ is used to update the momentum {\it queue}. 

In Eq.~\eqref{eq:supcon}, $x_{i}$ is the representation for image $X_{i}$ in $B_{v1}$ obtained by the encoder of {query network}. The transform $T(\cdot)$ also belongs to the {query network}. We write
\begin{align*}
A(i) ={}& \{z_{k} \in queue \cup Z_{v1} \cup Z_{v2}\} \backslash \{ z_{k} \in Z_{v1}: k=i\}, \\
P(i) ={}& \{z_{k} \in A(i): y_{k} = y_{i}\}.
\end{align*}

In implementation, the loss is usually scaled by $\cfrac{1}{|P(i)|}$ and a temperature $\tau$ is applied like in Eq.~\eqref{eq:infonce}.
Different from self-supervised contrastive loss, which treats query and key as a positive pair if they are the data-augmented version of the same image, supervised contrastive loss treats them as one positive pair if they belong to the same class. 

\subsection{Theoretical Motivation}
{\bf Analysis of Supervised Contrastive Learning.}
Khosla~\cite{DBLP:conf/nips/KhoslaTWSTIMLK20} introduced supervised contrastive learning to encourage more compact representation. We observe that it is not directly applicable to long-tailed recognition.
As shown in Table~\ref{tab:motivation}, the performance significantly decreases compared with traditional supervised cross-entropy.
From an optimization point of view, supervised contrastive loss concentrates more on high-frequency classes than low-frequency ones, which is unfriendly for imbalanced learning.

\begin{table}[tb!]
	\small
	\centering
	\caption{\textbf{Supervised contrastive learning is more sensitive to data imbalance.} Top-1 accuracy (\%) on ImageNet-LT with ResNet-50 is reported. Implementation details are in supplementary file. ``\dag" represents model is trained with PaCo loss without center learning rebalance.}
	\vspace{0pt}
	{
		\begin{tabular}{lcccc}
			\toprule
			Method &Many &Medium &Few &All \\
			\midrule
			Cross-Entropy &67.5 &42.6 &13.7 &48.4\\
			SupCon        &53.4 &2.9 &0 &22.0\\
			PaCo (ours) \dag    &69.6 &45.8 &16.0 &51.0\\
			\bottomrule
		\end{tabular}
	}
	\label{tab:motivation}
 \vspace{-0.1in}
\end{table}

\begin{theorem}
	\label{thm:01_supcon}
	\normalfont{(Optimal value for supervised contrastive learning).}
	When supervised contrastive loss converges, the optimal value for the probability that two samples are a true positive pair with label $y$ is $\cfrac{1}{K_{y}}$, where, $q(y)$ is the frequency of class $y$ over the whole dataset, {\it queue} is the momentum  {\it queue} in MoCo \cite{DBLP:conf/cvpr/He0WXG20, DBLP:journals/corr/abs-2003-04297} and $K_{y} \approx \mathtt{length}(queue) \cdot q(y)$.
\end{theorem}    

\vspace{+0.1in}
\noindent {\bf Interpretation.} As indicated by Remark \ref{thm:01_supcon}, high-frequency classes have a higher lower bound of loss value and contribute much more importance than low-frequency classes in training. Thus the training process can be dominated by high-frequency classes. To handle this issue, we introduce a set of parametric class-wise learnable centers for rebalancing in contrastive learning.

\subsection{Rebalance in Contrastive Learning}
\label{Sec:PaCo}
As described in Fig.~\ref{fig:paco_psum} (a), we introduce a set of parametric class-wise learnable centers $\mathbf{C} =\{c_{1}, c_{2},..., c_{n} \}$ into the original supervised contrastive learning, and called this new form {\bf Parametric Contrastive Learning (PaCo)}. Correspondingly, the loss function is changed to
\begin{equation}
\mathcal{L}_{i} \!=\! \sum_{ z_{+} \in P(i) \cup \{c_{y}\} } \!-w(z_{+}) \log \frac{ \exp({z_{+} \cdot T(x_{i})}) }{\sum_{z_{k} \in A(i) \cup \mathbf{C}} \exp({z_{k} \cdot T(x_{i})})},\label{eq:PaCo}
\end{equation}
\text{where}
\begin{equation}
w(z_{+})\!=\!\left\{
\begin{array}{lr}
\alpha, \!\quad\! z_{+} \!\in\! P(i) &  \\
1.0, \!\quad\! z_{+} \!\in\! \{c_{y} \} & \\
\end{array}
\right.
\nonumber
\end{equation}
and
\begin{equation}
z \cdot T(x_{i})\!=\!\left\{
\begin{array}{lr}
z \cdot \mathcal{G}(x_{i}), \!\quad\! z \!\in\! A(i) &  \\
z \cdot \mathcal{F}(x_{i}), \!\quad\! z \!\in\! \mathbf{C}. & \\
\end{array}
\right.
\nonumber
\end{equation}

Following Chen~\cite{DBLP:journals/corr/abs-2003-04297}, the transform $\mathcal{G}(\cdot)$ is a two-layer MLP while $\mathcal{F}(\cdot)$ is the identity mapping, {\textit i.e.}, $\mathcal{F}(x)=x$. $\alpha$ is one hyper-parameter in (0,1). $P(i)$ and $A(i)$ are the same with supervised contrastive learning in Eq.~\eqref{eq:supcon}. In implementation, the loss is scaled by $\cfrac{1}{\sum_{ z_{+} \in P(i) \cup \{ c_{y} \} } w(z_{+})}$ and a temperature $\tau$ is applied like in Eq.~\eqref{eq:supcon}.

\begin{theorem}
	\label{thm:02_pcl}
	\normalfont{(Optimal value for parametric contrastive learning)~~}
	When parametric contrastive loss converges, the optimal value for the probability that two samples are a true positive pair with label $y$ is $\cfrac{\alpha}{1+\alpha \cdot K_{y}}$ and the optimal value for the probability that a sample is closest to its corresponding center $c_{y}$ among $\mathbf{C}$ is $\cfrac{1}{1+\alpha \cdot K_{y}}$, where $q(y)$ is the frequency of class $y$ over the whole dataset, {\it queue} is the momentum  {queue} in MoCo \cite{DBLP:conf/cvpr/He0WXG20, DBLP:journals/corr/abs-2003-04297} and $K_{y} \approx \mathtt{length}(queue) \cdot q(y)$.
\end{theorem}  

\vspace{+0.1in}
\noindent {\bf Interpretation.} Suppose the most frequent class $y_{h}$ has $K_{y_{h}} \approx q(y_{h}) \cdot \mathtt{length}(queue)$ and the least frequent class $y_{t}$ has $K_{y_{t}} \approx q(y_{t}) \cdot \mathtt{length}(queue)$. As indicated by Remarks \ref{thm:02_pcl} and \ref{thm:01_supcon}, the optimal value for the probability that two samples are a true positive pair varying from the most frequent class to the least one is rebalanced from $\cfrac{1}{K_{y_{h}}}$ $\longrightarrow$ $\cfrac{1}{K_{y_{t}}}$ to $\cfrac{1}{\frac{1}{\alpha} + K_{y_{h}}}$ $\longrightarrow$ $\cfrac{1}{\frac{1}{\alpha} + K_{y_{t}}}$. The smaller $\alpha$, the more uniform the optimal value from the most frequent class to the least one is, friendly to low- frequency classes learning. 

However, when $\alpha$ decreases, the intensity of contrast among samples will be weaker, the intensity of contrast between samples and centers will be stronger. The whole loss becomes closer to supervised cross-entropy. To make good use of contrastive learning and rebalance at the same time, we observe that $\alpha$=0.05 is a reasonable choice.

\subsection{PaCo under balanced setting}
\label{sec:PaCo_balance}
For balanced datasets, all classes have the same frequency, {\it i.e.}, $q^{*}$=$q(y_{i})$=$q(y_{j})$ and $K^{*}$=$K_{y_{i}}$=$K_{y_{j}}$ for any class $y_{i}$ and class $y_{j}$. In this case, PaCo reduces to an improved version of multi-task with weighted sum of supervised cross-entropy loss and supervised contrastive loss. The connection between PaCo and multi-task is
\begin{equation*}
ExpSum = \sum_{c_{k} \in \mathbf{C}} \exp(c_{k} \!\cdot\! \mathcal{F}(x_{i})) + \sum_{z_{k} \in A(i)} \exp(z_{k} \cdot \mathcal{G}(x_{i})).
\end{equation*}

We also write the PaCo loss as
\begin{align*}
\mathcal{L}_{i} 
={}& \sum_{ z_{+} \in P(i) \cup \{c_{y}\} } \!-w(z_{+}) \log \frac{ \exp({z_{+} \cdot T(x_{i})}) }{\sum_{z_{k} \in A(i) \cup \mathbf{C}} \exp({z_{k} \cdot T(x_{i})})} \\
={}& -\log \frac{\exp(c_{y} \cdot \mathcal{F}(x_{i}) )}{ExpSum} - \alpha\sum_{z_{+} \in P(i) }\log \frac{\exp(z_{+} \cdot \mathcal{G}(x_{i}) )}{ExpSum}\\
={}& \mathcal{L}_{sup} \!+\! \alpha  \mathcal{L}_{supcon} \!-\! \left( \log P_{sup} \!+\! \alpha K^{*}  \log P_{supcon} \right)  \\
={}& \mathcal{L}_{sup} \!+\! \alpha \mathcal{L}_{supcon} \!-\! \left( \log P_{sup} \!+\! \alpha K^{*}  \log (1-P_{sup}) \right), \\
\notag
\end{align*}
\begin{equation}
\text{where}\quad \  \left\{
\begin{aligned}
P_{sup} &= \frac{\sum_{c_{k} \in \mathbf{C}} \exp(c_{k} \!\cdot\! \mathcal{F}(x_{i}))}{ExpSum}; \\
P_{supcon} &= \frac{\sum_{z_{k} \in A(i)} \exp(z_{k} \cdot \mathcal{G}(x_{i}))}{ExpSum}. \label{eq:pcl_multitask} \\
\end{aligned}
\right.
\end{equation}

Multi-task learning combines supervised cross-entropy loss and supervised contrastive loss with a fixed weighted scalar. When these two losses conflict, the training can suffer from slower or sub-optimization. Our PaCo contrarily adjust the intensity of supervised cross-entropy loss and supervised contrastive loss in an adaptive way and potentially avoids conflict as analyzed in the following.

\subsubsection{Analysis of PaCo under balanced setting}
As indicated by Eq.~\eqref{eq:pcl_multitask}, compared with multi-task, PaCo has an additional loss item:
\begin{equation}
\mathcal{L}_{extra}=-\log(P_{sup}) - \alpha K^{*} \log (1-P_{sup}). \label{eq:PaCo_extra}
\end{equation}

Here, we take full ImageNet as an example, {\it i.e.}, $q^{*}=0.001, \mathtt{length}(queue)=8192, \alpha=0.05, \alpha K^{*}=0.41$.
Then the function curve for $\mathcal{L}_{extra}$ is shown in Fig.~\ref{fig:paco_psum} (b).
With $P_{sup}$ increases from 0 to 1.0, the function value decreases until $P_{sup}=0.71$ and then goes up, which implies $\mathcal{L}_{extra}$ obtains the smallest loss value when $P_{sup}=0.71$. Note that, when the whole PaCo loss in Eq.~\eqref{eq:PaCo} achieves the optimal solution, $P_{sup}=0.71$ still establishes as demonstrated by Remark \ref{thm:02_pcl}. With $P_{sup}$ increases in the training, we analyze how does it affect the intensity of supervised contrastive loss and supervised cross-entropy loss in the following.

\vspace{+0.1in}
\noindent {bf Adaptive weighting between $\mathcal{L}_{sup}$ and $\mathcal{L}_{supcon}$.} 
Note that the optimal value for the probability that two samples are a true positive pair with label $y$ is 0.035 as indicated by Remark \ref{thm:02_pcl}. We suppose $p_{l}$, $p_{h}$ $\in$ (0, 0.71) and $p_{l}$ $\textless$ $p_{h}$. To achieve the optimal value, when $P_{sup}$=$p$, the supervised contrastive loss value $\mathcal{L}_{supcon}$ must decrease as in Eq.~\eqref{eq:PaCo_adaptive}.

\begin{equation}
\begin{aligned}
\mathcal{L}_{supcon} 
={}& -\sum_{z_{+} \in P(i)} \log \frac{\exp(z_{+} \cdot \mathcal{G}(x_{i}) )}{\sum_{z_{k} \in A(i)} \exp(z_{k} \cdot \mathcal{G}(x_{i})) } \\
={}& -\sum_{z_{+} \in P(i)} \log \frac{ \frac{\exp(z_{+} \cdot \mathcal{G}(x_{i}) )}{ExpSum}}{\frac{\sum_{z_{k} \in A(i)} \exp(z_{k} \cdot \mathcal{G}(x_{i}))}{ExpSum}} \\
={}& -K^{*} \log \frac{0.035}{1-p}. \label{eq:PaCo_adaptive} 
\end{aligned}
\end{equation}

Here $P_{sup}$ increases from $p_{l}$ to $p_{h}$, $\mathcal{L}_{supcon}$ must decrease to a much smaller loss value to achieve the optimal solution, which implies the need to make two different class samples much more discriminative, {\it i.e.}, increasing inter-class margins, and thus the intensity of supervised contrastive loss will enlarge.

An intuition is that as $P_{sup}$ increases, more samples are pulled together with their corresponding centers.
Along with stronger intensity of supervised contrastive loss at that time, it is more likely to push hard examples close to those samples that are already around right centers.

\subsection{Center Learning Rebalance}
PaCo balances the contrastive learning (for moderating contrast among samples). However the center learning also needs to be balanced, which has been explored in \cite{DBLP:journals/nn/BudaMM18,DBLP:conf/cvpr/HuangLLT16,DBLP:conf/cvpr/CuiJLSB19, he2009learning,chawla2002smote, shen2016relay, DBLP:conf/nips/RenYSMZYL20, DBLP:conf/iclr/KangXRYGFK20, DBLP:journals/corr/abs-2010-01809, DBLP:journals/corr/abs-2101-10633, DBLP:conf/nips/TangHZ20,duggal2020elf, Zhong_2021_CVPR}. We incorporate Balanced Softmax \cite{DBLP:conf/nips/RenYSMZYL20} into the center learning. Then the PaCo loss is changed from Eq.~\eqref{eq:PaCo} to what follows:
\begin{equation}
\mathcal{L}_{i} \!=\! \sum_{ z_{+} \in P(i) \cup \{c_{y}\} } \!-w(z_{+}) \log \frac{ \psi(z_{+}, T(x_{i})) }{\sum_{z_{k} \in A(i) \cup \mathbf{C}} \psi(z_{k}, T(x_{i}))},\label{eq:center_rebalance}
\end{equation}
\text{where}
\begin{equation}
\psi(z_{k}, T(x_{i}))\!=\!\left\{
\begin{array}{ll}
\exp(z_{k} \cdot \mathcal{G}(x_{i})), & z_{k} \!\in\! A(i);   \\
\exp(z_{k} \cdot \mathcal{F}(x_{i})) \cdot q(y_{k}), & z_{k} \!\in\! \mathbf{C}. \\
\end{array}
\right.
\nonumber
\end{equation}
We emphasize that Balanced Softmax is only a practical remedy for center learning rebalance. The theoretical analysis remains as a future work.

\subsection{Generalized Parametric Contrastive Learning}
\label{sec:gpaco}
MoCo \cite{DBLP:conf/cvpr/He0WXG20} and MoCo v2 \cite{DBLP:journals/corr/abs-2003-04297} use a momentum encoder and a queue to allow small mini-batch training and achieve even better performance than SimCLR \cite{DBLP:conf/icml/ChenK0H20} which requires large mini-batch training, while Chen et al. \cite{chen2021exploring} claims that it is "stop-gradient" rather than momentum encoder that avoid collapsing solutions in self-supervised learning.

We delve deeper to explore how each component in PaCo affects performance by examing the augmentation strategy, formulation of GPaCo/PaCo, the momentum encoder, and queue size in Sec. \ref{sec:ablation_study}. Interestingly, we observe that the derived Generalized Parametric Contrastive Learning (GPaCo) by removing the momentum encoder boosts performance on imbalanced and balanced data again, indicating that the strong regularization effects from the momentum encoder can be harmful to the model performance in supervised leanring.

In addition to long-tailed recognition, GPaCo models enjoy better generalization ability across CNNs (ResNets) to vision transformers (ViT models) on full ImageNet and CIFAR, which is demonstrated in Sec. \ref{sec:full_imagenet_cifar} and Sec. \ref{sec:robustness}.
Besides image classification, GPaCo shows its great generality on other tasks, {\textit e.g.}, out-of-distribution robustness, and semantic segmentation. 

To demonstrate the advantages of GPaCo on out-of-distribution data, we load MAE \cite{he2022masked} pre-trained weights and then fine-tune on full ImageNet with same training strategy as in \cite{he2022masked}. Compared with MAE baselines, GPaCo models achieve much stronger robustness across different benchmarks, which is discussed in Sec. \ref{sec:robustness}.

For semantic segmentation task, GPaCo treats each pixel as an example. However, huge number of pixels make it unpractical to directly apply GPaCo to pixel classification. To reduce computational cost and GPU memory, we randomly sample 8192 pixels from down-sampled pixel features. These selected pixel features go through a 3-layer mlp transform and then are fed into GPaCo, which is adopted as an auxiliary loss for training optimization. GPaCo models on semantic segmentation show obvious improvements compared with baselines across 4 most popular datasets in semantic segmentation community. More details are introduced in Sec. \ref{sec:semantic_seg}.

%% file: doc/exp.tex
\section{Experiments}
\subsection{Long-tailed Recognition}
We follow the common evaluation protocol \cite{Liu_2019_CVPR, DBLP:journals/corr/abs-2101-10633,DBLP:conf/iclr/KangXRYGFK20} in long-tailed recognition -- that is, training models on the long-tailed source label distribution and evaluating their performance on the uniform target label distribution.
We conduct experiments on long-tailed version of CIFAR-100 \cite{cb-focal, DBLP:conf/nips/CaoWGAM19}, Places~\cite{Liu_2019_CVPR}, ImageNet \cite{Liu_2019_CVPR} and iNaturalist 2018 \cite{van2018inaturalist} datasets.

\begin{table}[t]
	\centering
	\caption{\textbf{Top-1 accuracy on ImageNet-LT for different backbone architectures.}  ``\dag"~denotes models are trained with RandAugment \cite{DBLP:conf/nips/CubukZS020} in 400 epochs. More comparisons with RIDE \cite{DBLP:journals/corr/abs-2010-01809} are in Fig.~\ref{fig:comparison}.}
	\label{tab:imagenet_main}
	{
		\begin{tabular}{lc@{\ \ }c@{\ \ }c} 
			\toprule
			Method & ResNet-50 & ResNeXt-50 & ResNeXt-101 \\
			\midrule
			CE(baseline)                &41.6 &44.4 &44.8 \\
			Decouple-cRT                &47.3 &49.6 &49.4 \\
			Decouple-$\tau$-norm        &46.7 &49.4 &49.6 \\
			De-confound-TDE    &51.7 &51.8 &53.3 \\
			ResLT              &-    &52.9 &54.1 \\
			MiSLAS             &52.7 &-    & -   \\
			\midrule
			Decouple-$\tau$-norm \dag     &54.5 &56.0 &57.9 \\
			Balanced Softmax \dag         &55.0 &56.2 &58.0 \\
			PaCo\dag                      &57.0 &58.2 &60.0 \\
			GPaCo\dag                     &\textbf{58.5} &\textbf{58.9} &\textbf{60.8} \\
			\bottomrule
		\end{tabular}
	}
	\vspace{-0.15in}
\end{table}

\vspace{+0.1in}
\noindent {\bf CIFAR-100-LT datasets.}
We use the long-tailed version of CIFAR datasets with the same setting as those used in \cite{cao2019learning, zhou2019bbn, cb-focal}. They control the degrees of data imbalance with an imbalance factor $\beta$. $\beta$= $\frac{N_{max}}{N_{min}}$ where $N_{max}$ and $N_{min}$ are the numbers of training samples for the most and least frequent classes respectively. Following \cite{zhou2019bbn}, we conduct experiments with imbalance factors 100, 50, and 10.

\vspace{+0.1in}
\noindent {\bf ImageNet-LT and Places-LT.}
ImageNet-LT and Places-LT were proposed in \cite{Liu_2019_CVPR}. ImageNet-LT is a long-tailed version of ImageNet dataset \cite{imagenet} by sampling a subset following the Pareto distribution with power value $\alpha$=6. It contains 115.8K images from 1,000 categories, with class cardinality ranging from 5 to 1,280. Places-LT is a long-tailed version of the large-scale scene classification dataset Places \cite{zhou2017places}. It consists of 184.5K images from 365 categories with class cardinality ranging from 5 to 4,980.

\vspace{+0.1in}
\noindent{ \bf iNaturalist 2018.}
The iNaturalist 2018 \cite{van2018inaturalist} is one species classification dataset, which is on a large scale and suffers from extremely imbalanced label distribution. It is composed of 437.5K images from 8,142 categories. In addition to the extreme imbalance, the iNaturalist 2018 dataset also confronts the fine-grained problem \cite{wei2019piecewise}.

\vspace{+0.1in}
\noindent{\bf Implementation details.}
For image classification on ImageNet-LT, we used ResNet-50, ResNeXt-50-32x4d, and ResNeXt-101-32x4d as our backbones for experiments. For iNaturalist 2018, we conduct experiments with ResNet-50 and ResNet-152. All models were trained using SGD optimizer with momentum $\mu = 0.9$. 

Contrastive learning benefits from longer training compared with traditional supervised learning with cross-entropy as Chen~\cite{DBLP:conf/icml/ChenK0H20} concluded, which is also validated by previous work of \cite{DBLP:conf/icml/ChenK0H20, DBLP:conf/cvpr/He0WXG20, DBLP:journals/corr/abs-2003-04297, DBLP:conf/nips/GrillSATRBDPGAP20, DBLP:conf/nips/CaronMMGBJ20}.
MoCo \cite{DBLP:conf/cvpr/He0WXG20, DBLP:journals/corr/abs-2003-04297}, BYOL \cite{DBLP:conf/nips/GrillSATRBDPGAP20} and SWAV \cite{DBLP:conf/nips/CaronMMGBJ20} train 800 epochs for model convergence. Supervised contrastive learning \cite{DBLP:conf/nips/KhoslaTWSTIMLK20} trains 350 epochs for feature learning and another 350 epochs for classifier learning. 

Following MoCo \cite{DBLP:conf/cvpr/He0WXG20, DBLP:journals/corr/abs-2003-04297}, when we train models with GPaCo/PaCo, the learning rate decays by a cosine scheduler from 0.04/0.02 to 0 with batch size 128 on 4 GPUs in 400 epochs. The temperature is set to 0.2. $\alpha$ is 0.05.
For a fair comparison, we re-implement baselines with the same training time and RandAugment \cite{DBLP:conf/nips/CubukZS020} for recent state-of-the-arts of Decouple \cite{DBLP:conf/iclr/KangXRYGFK20}, Balanced Softmax \cite{DBLP:conf/nips/RenYSMZYL20} and RIDE~\cite{DBLP:journals/corr/abs-2010-01809}. Especially, for RIDE, based on model ensemble, we compare with it under comparable inference latency in Fig.~\ref{fig:comparison} (a).

For Places-LT, following previous setting \cite{Liu_2019_CVPR, DBLP:journals/corr/abs-2101-10633},
we choose ResNet-152 as the backbone network, pre-train it on the full ImageNet-2012 dataset (provided by torchvision), and finely tune it for 30 epochs on Places-LT. The learning rate decays by a cosine scheduler from 0.02 to 0 with batch size 128. The temperature is set to 0.2. $\alpha$ is 0.02/0.05 for GPaCo and PaCo. For CIFAR-100-LT, we strictly follow the setting of \cite{DBLP:conf/nips/RenYSMZYL20} for fair comparison. A smaller temperature of 0.07 and $\alpha= 0.01$ are adopted.

\vspace{+0.1in}
\noindent{ \bf Comparison on ImageNet-LT.}
Table~\ref{tab:imagenet_main} shows extensive experimental results for comparison with recent SOTA methods. We observe that Balanced Softmax \cite{DBLP:conf/nips/RenYSMZYL20} still achieves comparable results with Decouple \cite{DBLP:conf/iclr/KangXRYGFK20} across various backbones under such strong training setting on ImageNet-LT, consistent with what is claimed in the original paper. For RIDE that is based on model ensemble, we analyze the real inference speed by calculating inference time with a batch of 64 images on Nvidia GeForce 2080Ti GPU. 

We observe RIDEResNet with 3 experts even has higher inference latency than a standard ResNeXt-50-32x4d (\textbf{15.3ms vs 13.1ms}); RIDEResNeXt with 3 experts yields higher inference latency than a standard ResNeXt-101-32x4d (\textbf{26ms vs 25ms}). This result is in accordance with the conclusion that network fragmentation reduces the degree of parallelism and thus decreases efficiency in \cite{DBLP:conf/eccv/MaZZS18, DBLP:conf/iccv/CuiCLLSJ19}. For fair comparison, we do not apply knowledge distillation tricks for all these methods. As shown in Fig.~\ref{fig:comparison} (a) and Table~\ref{tab:imagenet_main}, under comparable inference latency, GPaCo/PaCo significantly surpasses these baselines. 

\begin{table}[t]
	\centering
	\caption{\textbf{Performance on Places-LT~\cite{Liu_2019_CVPR}, starting from an ImageNet pre-trained ResNet-152 provided by torchvision.} ``\dag" denotes the model trained with RandAugment \cite{DBLP:conf/nips/CubukZS020}.}
	\label{tab:place_main} 
	{
		\begin{tabular}{lc@{\ \ }c@{\ \ }c@{\ \ }c}
			\toprule
			Method & Many & Medium & Few & \textbf{All}      \\
			\midrule
			CE(baseline)      & 45.7 & 27.3 & 8.2 & 30.2     \\
			OLTR              & 44.7 & 37.0 & 25.3 & 35.9    \\
			Decouple-$\tau$-norm & 37.8 & 40.7 & 31.8 & 37.9 \\
			Balanced Softmax     & 42.0 & 39.3 & 30.5 & 38.6 \\
			ResLT                & 39.8 & 43.6 & 31.4 & 39.8 \\
			MiSLAS               & 39.6 & 43.3 & 36.1 & 40.4 \\
			RIDE (2 experts)     & -    & -    & -    & -    \\ 
			\midrule
			PaCo                  & 37.5 & 47.2 & 33.9 & 41.2 \\
			PaCo \dag             & 36.1 & 47.9 & 35.3 & 41.2 \\      
			GPaCo                 & 39.5 & 47.2 & 33.0 &\textbf{41.7} \\
			\bottomrule
		\end{tabular}
	}
	\vspace{-0.15in}
\end{table}

\begin{table*}
	\centering
	\setlength{\tabcolsep}{7.5pt}
	\caption{\textbf{Evaluation with ViT models\cite{dosovitskiy2020image} on full ImageNet and out-of-distribution data.} GPaCo models enjoy better generalization ability and stronger robustness compared with MAE models. We use their official open-source code for reproduing results of MAE models with 8 Nvidia GeForce RTX3090 GPUs. Top-1 accuracy are reported for full ImageNet, ImageNet-R, and ImageNet-S. mCE and rel. mCE are reported for ImageNet-C.}
	\label{tab:full_imagenet_vit}
	{
		\begin{tabular}{lcccccc}
			\toprule
			Method &Model &Full ImageNet ($\uparrow$) &ImageNet-C (mCE $\downarrow$) &ImageNet-C (rel. mCE $\downarrow$) & ImageNet-R ($\uparrow$) &ImageNet-S ($\uparrow$) \\
			\midrule
			\multicolumn{7}{c}{w/ Mixup and CutMix} \\
			\midrule
			MAE    &ViT-B &83.6 &39.1 &49.9 &49.9 &36.1 \\
			MAE    &ViT-L &85.7 &32.4 &41.4 &60.3 &45.5 \\
			\midrule
			\multicolumn{7}{c}{w/o Mixup and CutMix} \\
			\midrule
			SupCon       &ResNet-50 &78.7 &67.2 &94.6 &-     &-    \\
			GPaCo        &ResNet-50 &\textbf{79.7} &\textbf{50.9} &\textbf{64.4} &\textbf{41.1} &\textbf{30.9} \\
			\midrule
			MAE    &ViT-B &82.5 &44.9 &56.9 &44.9 &32.9 \\
			MAE    &ViT-L &85.2 &36.4 &46.3 &55.3 &42.6 \\
			GPaCo        &ViT-B  &\textbf{84.0} &\textbf{37.2} &\textbf{47.3} &\textbf{51.7} &\textbf{39.4} \\
			GPaCo        &ViT-L  &\textbf{86.0} &\textbf{30.7} &\textbf{39.0} &\textbf{60.3} &\textbf{48.3} \\
			\bottomrule
		\end{tabular}
	}
\end{table*}

\begin{table}
	\centering
	\setlength{\tabcolsep}{5pt}
	\caption{\textbf{Top-1 accuracy on full ImageNet with ResNets.} ``$\star$" denotes supervised contrastive learning with additional operation of image warping before Gaussian blur.}
	\label{tab:full_imagenet}
	{
		\begin{tabular}{lccc}
			\toprule
			Method & Model & augmentation &Top-1 Acc \\
			\midrule
			Supcon     &ResNet-50  &SimAugment $\star$   &77.9 \\
			Supcon     &ResNet-50  &RandAugment          &78.4 \\
			Supcon     &ResNet-101 &StackedRandAugment   &80.2 \\
			\midrule
			multi-task &RandAugment         &ResNet-50 &78.1 \\
			\midrule
			PaCo        &ResNet-50  &SimAugment           &\textbf{78.7} \\
			PaCo        &ResNet-50  &RandAugment          &\textbf{79.3} \\
			PaCo        &ResNet-101 &StackedRandAugment   &\textbf{80.9} \\
			PaCo        &ResNet-200 &StackedRandAugment   &\textbf{81.8} \\
			\midrule
			GPaCo        &ResNet-50  &RandAugment         &\textbf{79.5} \\
			GPaCo        &ResNet-50  &StackedRandAugment  &\textbf{79.7} \\
			\bottomrule
		\end{tabular}
	}
\end{table}

\vspace{+0.1in}
\noindent{\bf Comparison on Places-LT.}
The experimental results on Places-LT are summarized in Table~\ref{tab:place_main}. Due to the architecture change of RIDE, it is not applicable to load the publicly pre-trained model on full ImageNet, while GPaCo/PaCo is more flexible because the network architecture is the same as those of \cite{Liu_2019_CVPR, DBLP:journals/corr/abs-2101-10633}. Under fair training setting by finely tuning 30 epochs without RandAugment, PaCo surpasses SOTA Balanced Softmax by 2.6\%. GPaCo even achieves 41.7\% after removing the momentum encoder. An interesting observation is that RandAugment has little effect on the Places-LT dataset. A similar phenomenon can be observed on the iNaturalist 2018 dataset. More evaluation numbers are in the supplementary file. They can be intuitively understood since RandAugment is designed for ImageNet classification, which inspires us to explore general augmentations across different domains.

\vspace{+0.1in}
\noindent{\bf Comparison on iNaturalist 2018.}
Table~\ref{tab:inat_main} lists experimental results on iNaturalist 2018. Under fair training setting, PaCo consistently surpasses recent SOTA methods of Decouple, Balanced Softmax and RIDE. Our method is 1.4\% higher than Balanced Softmax. We also apply PaCo on large ResNet-152 architecture. And the performance boosts to \textbf{75.3\%} top-1 accuracy. Surprisingly, With GPaCo, a ResNet-50 model achieves \textbf{75.4\%} top-1 accuracy, implying that the momentum encoder for PaCo can hurt optimization in training.
Note that we only transfer the hyper-parameters of PaCo on ImageNet-LT to iNaturalist 2018 without any change. Tuning hyper-parameters for GPaCo/PaCo will bring further improvement.

\vspace{+0.1in}
\noindent{\bf Comparison on CIFAR-100-LT.}
The experimental results on CIFAR-100-LT are listed in Table~\ref{tab:cifar100_main}. For the CIFAR-100-LT dataset, we mainly compare with the SOTA method Balanced Softmax \cite{DBLP:conf/nips/RenYSMZYL20} with the same training setting where Cutout \cite{DBLP:journals/corr/abs-1708-04552} and AutoAugment \cite{DBLP:conf/cvpr/CubukZMVL19} are used in training. As shown in Table~\ref{tab:cifar100_main}, PaCo consistently outperforms Balanced Softmax across different imbalance factors with such a strong setting. Specifically, PaCo surpasses Balanced Softmax by 1.2\%, 1.8\% and 1.2\% under imbalance factor 100, 50 and 10 respectively, which testify the effectiveness of our PaCo method. GPaCo models show better performance than PaCo models.

\subsection{Full ImageNet and CIFAR Recognition}
\label{sec:full_imagenet_cifar}
As analyzed in Section \ref{sec:PaCo_balance}, for balanced datasets, PaCo reduces to an improved version of multi-task learning, which adaptively adjusts the intensity of supervised cross-entropy loss and supervised contrastive loss. To verify the effectiveness of PaCo under this balanced setting, we conduct experiments on full ImageNet and full CIFAR. They are indicative to compare GPaCo/PaCo with supervised contrastive learning (Supcon) \cite{DBLP:conf/nips/KhoslaTWSTIMLK20}. Note that, under full ImageNet and CIFAR, we remove the rebalance in center learning, {\it i.e.}, Balanced Softmax, for fair comparisons.

\begin{table}[t]
	\centering
	\setlength{\tabcolsep}{30pt}
	\caption{\textbf{Top-1 accuracy over all classes on iNaturalist 2018 with ResNet-50.} Knowledge distillation is not applied for fair comparison. We compare with RIDE under comparable inference latency. ``\dag"~denotes models trained with RandAugment \cite{DBLP:conf/nips/CubukZS020} in 400 epochs.}
	\vspace{+0.03in}
	\label{tab:inat_main}
	{
		\begin{tabular}{lc}
			\toprule
			Method & Top-1 Acc \\
			\midrule
			LDAM+DRW                   & 68.0 \\
			Decouple-LWS               & 69.5 \\
			BBN                        & 69.6 \\
			ResLT                      & 70.2 \\
			MiSLAS                     & 71.6 \\
			\midrule
			RIDE (2 experts) \dag         & 69.5 \\
			Decouple-$\tau$-norm \dag     & 71.5 \\
			Balanced Softmax \dag         & 71.8 \\
			PaCo \dag                     & 73.2 \\
			GPaCo \dag                    & \textbf{75.4} \\
			\bottomrule
		\end{tabular}
	}
\end{table}

\vspace{+0.1in}
\noindent{\bf Full ImageNet.}
In the implementation, we transfer hyper-parameters of GPaCo/PaCo on ImageNet-LT to full ImageNet without modification. SGD optimizer with momentum $\mu = 0.9$ is used. $\alpha$=0.05, temperature is 0.2 and queue size is 8,192. 
For multi-task training, the supervised contrastive loss is an additional regularization and the loss weight is extensively explored from 0.1 to 1.0 in Sec. \ref{sec:ablation_study}. The same data augmentation strategy is applied as in GPaCo/PaCo, which is discussed in Section~\ref{Sec:data_augmentation}. 

The experimental results are summarized in Table~\ref{tab:full_imagenet}. With SimAugment, our ResNet-50 PaCo model achieves 78.7\% top-1 accuracy, which outperforms supervised contrastive learning model by 0.8\%. Equipped with strong augmentation, {\it i.e.}, RandAugment \cite{DBLP:conf/nips/CubukZS020}, the performance further improves to 79.3\%. ResNet-101/200 trained with PaCo consistently surpass supervised contrastive learning. 
Replacing PaCo with GPaCo, The ResNet-50 model boosts to \textbf{79.7\%} top-1 accuracy.

\vspace{+0.1in}
\noindent{\bf Full CIFAR-100.}
For CIFAR implementation, we follow supervised contrastive learning and train ResNet-50 with only the SimAugment. Compared with full ImageNet, we adopt a smaller temperature of 0.07, $\alpha=0.01$ and batch size 256 with learning rate 0.1. As shown in Table~\ref{tab:full_cifar}, on CIFAR-100, PaCo outperforms supervised contrastive learning by 2.6\%, which validates the advantages of PaCo. Again, the GPaCo model achieves \textbf{80.3\%} top-1 accuracy, significantly surpassing supervised contrastive learning by \textbf{3.8\%}. 
Note that, following \cite{DBLP:conf/iccv/CuiCLLSJ19}, we use a weight-decay of 5e-4.

\begin{table}[t]
	\centering
	\caption{\textbf{Transfering GPaCo to the semantic segmentation.} }
	\label{tab:ade20k_smseg}
	{
		\begin{tabular}{lccc}
			\toprule
			Method &Backbone &mIoU (s.s.) &mIoU (m.s.) \\
			\midrule
			\multicolumn{4}{c}{\textbf{ADE20K Dataset}} \\
			\midrule
			\multirow{3}{*}{UperNet}  &Swin-T &44.5 &45.8\\
			                          &Swin-B &50.0 &51.7\\
			                          &Swin-L &52.0 &53.5\\
			\midrule
			\multirow{3}{*}{UperNet w/ GPaCo}     &Swin-T &\textbf{45.4} &\textbf{46.8} \\
		                	                      &Swin-B &\textbf{51.6} &\textbf{53.2} \\
			                                      &Swin-L &\textbf{52.8} &\textbf{54.3} \\
			\midrule
			\multicolumn{4}{c}{\textbf{COCO-Stuff Dataset}} \\
			\midrule
			\multirow{2}{*}{deeplabv3+} &ResNet-50  &35.8 &36.8\\
			                            &ResNet-101 &38.1 &39.0\\
			\midrule
			\multirow{2}{*}{deeplabv3+ w/ GPaCo}     &ResNet-50  &\textbf{37.0} &\textbf{37.9} \\
		                	                         &ResNet-101 &\textbf{38.8} &\textbf{40.1} \\
		    \midrule
			\multicolumn{4}{c}{\textbf{PASCAL Context Dataset}} \\
			\midrule
			\multirow{2}{*}{deeplabv3+} &ResNet-50  &50.5 &52.1\\
			                            &ResNet-101 &52.9 &54.5\\
			\midrule
			\multirow{2}{*}{deeplabv3+ w/ GPaCo}     &ResNet-50  &\textbf{51.9} &\textbf{53.7} \\
		                	                         &ResNet-101 &\textbf{54.2} &\textbf{56.2} \\
		    \midrule
		    \multicolumn{4}{c}{\textbf{Cityscapes}} \\
			\midrule
			\multirow{3}{*}{deeplabv3+} &ResNet-18  &76.9 &78.8 \\
			                            &ResNet-50  &80.1 &81.1 \\
			                            &ResNet-101 &81.0 &82.0 \\
			\midrule
			\multirow{3}{*}{deeplabv3+ w/ GPaCo}     &ResNet-18  &\textbf{78.1} &\textbf{79.7} \\
			                                         &ResNet-50  &\textbf{80.8} &\textbf{82.0} \\
		                	                         &ResNet-101 &\textbf{81.4} &\textbf{82.1} \\
		                	                         
			\bottomrule
		\end{tabular}
	}
\end{table}

\subsection{Out-of-distribution Robustness}
\label{sec:robustness}
He et al. \cite{he2022masked} justifies that masked autoencoders are scalable vision learners. With such a simple pre-training strategy that masking random patches of the input image and reconstructing the missing pixels, the good representation is learned and well transffered to downstream tasks, {\textit i.e.}, full ImageNet classification, object detection, semantic segmentaiton, and robustness on out-of-distribution data. In this section, we verify that GPaCo can enhance model generalization ability and robustness on full ImageNet and its variants when compared with MAE \cite{he2022masked} models. 

\vspace{+0.1in}
\noindent {\bf Robustness Evaluation.} 
We extensively evaluate the model performance on out-of-distribution robustness using following benchmarks: 1) ImageNet-C \cite{hendrycks2018benchmarking}, with various common image corruptions; 2) ImageNet-R \cite{hendrycks2021many}, which contains natural renditions of ImageNet object classes with different textures and local image statistics; 3) ImageNet-Sketch \cite{wang2019learning}, which includes sketch images of the same ImageNet classes collected online.

\begin{table}[t]
	\setlength{\tabcolsep}{15pt}
	\centering
    \caption{\textbf{Top-1 accuracy on CIFAR-100-LT with different imbalance factors.} ``\dag" represents that models are trained in same setting.}
	{
		\begin{tabular}{l  c  c  c}
			\toprule
			Dataset & \multicolumn{3}{c}{CIFAR-100 LT} \\ 
			\midrule
			Imbalance factor &100 &50 & 10 \\
			\midrule
			Focal Loss  & 38.4 & 44.3 & 55.8 \\
			LDAM+DRW    & 42.0 & 46.6 & 58.7 \\
			BBN         & 42.6 & 47.0 & 59.1 \\
			Causal Norm & 44.1 & 50.3 & 59.6 \\
			ResLT       & 45.3 & 50.0 & 60.8 \\
			MiSLAS      & 47.0 & 52.3 & 63.2 \\
			\midrule
			
			Balanced Softmax \dag &50.8 &54.2 &63.0 \\
			PaCo \dag    &52.0 &56.0 &64.2 \\
			GPaCo \dag   &\textbf{52.3} &\textbf{56.4} &\textbf{65.4} \\
			\bottomrule
		\end{tabular}
	}
	\label{tab:cifar100_main}
\end{table}

\begin{table}[t]
	\setlength{\tabcolsep}{19pt}
	\centering
	\caption{\textbf{Top-1 accuracy on full CIFAR-100 with ResNet-50.}}
	\label{tab:full_cifar}
	{
		\begin{tabular}{lccc}
			\toprule
			Method & dataset &Top-1 Acc \\
			\midrule
			CE(baseline)    &CIFAR-100    &77.9 \\
			multi-task      &CIFAR-100    &78.0 \\
			\midrule
			Supcon          &CIFAR-100    &76.5 \\
			PaCo            &CIFAR-100    &79.1 \\
			GPaCo           &CIFAR-100    &\textbf{80.3} \\
			\bottomrule
		\end{tabular}
	}
\end{table}

\vspace{+0.1in}
\noindent {\bf Comparison with MAE Models \cite{he2022masked}.}
We summarize experimental results in Table~\ref{tab:full_imagenet_vit}. Top-1 accuracy is reported on full ImageNet, ImageNet-R, and ImageNet-S. Mean Corruption Error (mCE) and Relative Mean Corruption Error (rel. mCE) \cite{hendrycks2018benchmarking} are used for ImageNet-C. mCE is to measure absolute robustness to corruptions while rel. mCE is a better metric when we compare models with different top-1 accuracy.

Compared with MAE models, we achieve better performance on full ImageNet, surpassing them by \textbf{0.4} and \textbf{0.3} respectively for ViT-B \cite{dosovitskiy2020image} and ViT-L \cite{dosovitskiy2020image}. On ImageNet-C, ImageNet-R, and ImageNet-S, GPaCo models usually outperforms MAE models by \textbf{1.7 $\sim$ 3.3}, demonstrating much stronger robustness on out-of-distribution data.

\subsection{Semantic Segmentation}
\label{sec:semantic_seg}
In this section, we transfer GPaCo to the downstream task, {\textit i.e.}, semantic segmentation. In implementation, we use GPaCo as an auxiliary loss. Specifically, for each image, we randomly sample 8192 pixel features on the donwn-sampled feature map and feed them into GPaCo loss for training optimization. 

\vspace{+0.1in}
\noindent {\bf Datasets.}
ADE20K \cite{zhou2017scene} contains 22K densely annotated images with 150 fine-grained semantic concepts. The training and validation sets consist of 20K and 2K images, respectively.
COCO-Stuff \cite{caesar2018coco} includes 10K images from the COCO training set. The training and validation sets consist of 9K and 1K images, respectively. It covers 171 classes.
For PASCAL Context \cite{mottaghi2014role}, the subset of 59 frequent classes is used following previous work.
Cityscapes \cite{cordts2016cityscapes} consists of 19 classes, covering street scenes from 50 different cities.

\vspace{+0.1in}
\noindent {\bf Training and Evaluation.}
We implement our GPaCo in mmseg codebase \cite{mmseg2020}. On ADE20K \cite{zhou2017scene}, we follow \cite{liu2021swin} to experiment with Swin transformers and UperNet \cite{xiao2018unified}. On COCO-Stuff \cite{caesar2018coco}, Pascal Context \cite{mottaghi2014role} and Cityscapes \cite{cordts2016cityscapes}, we use ResNet-50/101 and deeplabv3+. 
Default training hyper-parameters in baselines are adopted in the GPaCo model training phase, {\textit i.e.}, the standard random scale jittering between 0.5 and 2.0, random horizontal flipping, random cropping, as well as random color jittering. We report both single scale (s.s.) and multi-scale (m.s.) evaluation results on validation data. For multi-scale inference, scales of 0.5, 0.75, 1.0, 1.25, 1.5, 1.75 are used.

\vspace{+0.1in}
\noindent {\bf Results.}
Experimental results are summarized in Table~\ref{tab:ade20k_smseg}. On ADE20K, obvious improvements are observed from Swin-T to Swin-L models when equipped with GPaCo, outperforming baselines by \textbf{1.0}, \textbf{1.5}, \textbf{0.8} mIoU respectively. On COCO-Stuff and PASCAL Context, GPaCo models significantly surpass baselines by around \textbf{1.1 $\sim$ 1.7} mIoU.

\begin{table}[t]
	\centering
	\setlength{\tabcolsep}{10pt}
	\caption{\textbf{Ablation on augmentation strategies for GPaCo/PaCo on ImageNet-LT with ResNet-50.}}
	\label{tab:ablation_strongaug}
	{
		\begin{tabular}{lccc}
			\toprule
			Methods  &View-1 & View-2 &Top-1 Acc \\
			\midrule
			PaCo      &SimAug  &SimAug   &55.0 \\
			PaCo      &RandAug &SimAug   &57.0 \\
			PaCo      &RandAug &RandAug  &56.5 \\
			\midrule
			GPaCo     &RandAug &SimAug        &57.9 \\
			GPaCo     &RandAug &RandAugStack  &58.5 \\
			\bottomrule
		\end{tabular}
	}
\end{table}

\begin{table}[t]
	\centering
	\setlength{\tabcolsep}{22pt}
	\caption{\textbf{Comparison with multi-task re-weighting baselines on ImageNet-LT with ResNet-50.} The re-weighting strategy is applied to the supervised contrastive loss. Models are all trained without RandAugment. }
	\label{tab:ablation_reweighting_supp}
	\vspace{0pt}
	{
		\begin{tabular}{lc}
			\toprule
			Method  &Top-1 Acc \\
			\midrule
			CE                              &48.4 \\
			multi-task (CE+Re-weighting)    &49.0 \\
			multi-task (CE+BalSfx)  &48.6 \\
			\midrule
			PaCo                &51.0 \\
			\bottomrule
		\end{tabular}
	}
\end{table}

\begin{table}[t]
	\centering
	\setlength{\tabcolsep}{8pt}
	\caption{\textbf{Comparison with multi-task re-weighting baselines that perform center learning rebalance on ImageNet-LT.} Models are all trained with RandAugment in 400 epochs.}
	\label{tab:ablation_center_rebalance_supp}
	\vspace{0pt}
	{
		\begin{tabular}{lccc}
			\toprule
			Method  &Backbone &Weight &Top-1 Acc \\
			\midrule
			multi-task (BalSfx \!+\! rw)      &ResNeXt-50 &0.05 &57.0  \\
			multi-task (BalSfx \!+\! rw)      &ResNeXt-50 &0.10 &57.1  \\
			multi-task (BalSfx \!+\! rw)      &ResNeXt-50 &0.20 &57.1  \\
			multi-task (BalSfx \!+\! rw)      &ResNeXt-50 &0.30 &57.0  \\
			multi-task (BalSfx \!+\! rw)      &ResNeXt-50 &0.50 &57.2  \\
			multi-task (BalSfx \!+\! rw)      &ResNeXt-50 &0.80 &57.2  \\
			multi-task (BalSfx \!+\! rw)      &ResNeXt-50 &1.00 &56.9  \\
			PaCo                &ResNeXt-50 &- &58.2 \\
			\midrule
			multi-task* (BalSfx \!+\! rw)     &ResNeXt-50 &0.50 &58.3  \\
			multi-task* (BalSfx \!+\! rw)     &ResNeXt-50 &0.80 &58.2  \\
			GPaCo               &ResNeXt-50 &- &58.9 \\
			\midrule
			multi-task* (BalSfx \!+\! rw)     &ResNet-50 &0.50 &57.8  \\
			GPaCo                             &ResNet-50 &-    &58.5  \\
			\midrule
			multi-task* (BalSfx \!+\! rw)     &ResNeXt-101 &0.50 &60.0  \\
			GPaCo                             &ResNeXt-101 &-    &60.8  \\
			\bottomrule
		\end{tabular}
	}
\end{table}

\begin{table}[t]
	\centering
	\setlength{\tabcolsep}{10pt}
	\caption{\textbf{Comparison with multi-task baselines on balanced data.}}
	\label{tab:ablation_multitask_cifar100}
	\resizebox{1.0\linewidth}{!}
	{
		\begin{tabular}{lccc}
			\toprule
			Method  &Backbone &Weight &Top-1 Acc \\
			\midrule
			\multicolumn{4}{c}{\textbf{Full CIFAR-100 Dataset}} \\
			\midrule
			multi-task (CE \!+\! Supcon)       &ResNet-50 &0.50 &78.0  \\
			PaCo                &ResNet-50 &- &79.1 \\
			\midrule
			multi-task* (CE \!+\! Supcon)      &ResNet-50 &0.10 &79.0  \\
			multi-task* (CE \!+\! Supcon)      &ResNet-50 &0.30 &79.1  \\
			multi-task* (CE \!+\! Supcon)      &ResNet-50 &0.50 &79.1  \\
			multi-task* (CE \!+\! Supcon)      &ResNet-50 &0.80 &78.9  \\
			multi-task* (CE \!+\! Supcon)      &ResNet-50 &1.00 &79.0  \\
			GPaCo               &ResNet-50 &- &80.3 \\
			\midrule
			\multicolumn{4}{c}{\textbf{Full ImageNet Dataset}} \\
			\midrule
			multi-task* (CE \!+\! Supcon)      &ViT-B &0.50 &83.4  \\
			GPaCo               &ViT-B &- &84.0 \\
			\bottomrule
		\end{tabular}
	}
\end{table}

\subsection{Ablation Study}
\label{sec:ablation_study}
\vspace{+0.1in}
\noindent{\bf Data augmentation strategy for PaCo \& GPaCo.}
\label{Sec:data_augmentation}
Data augmentation is the key for success of contrastive learning as indicated by Chen~\cite{DBLP:conf/icml/ChenK0H20}. For PaCo \& GPaCo, we also conduct ablation studies for different augmentation strategies. Several observations are intriguingly different from those of \cite{ouyang2016factors}.
We experiment with the following ways of data augmentation.
\begin{itemize}
	\item SimAug: an augmentation policy \cite{DBLP:conf/cvpr/He0WXG20, DBLP:journals/corr/abs-2003-04297} that applies random flips and color jitters followed by Gaussian blur.
	\item RandAug \cite{DBLP:conf/nips/CubukZS020}: A two stage augmentation policy that uses random parameters in place of parameters tuned by AutoAugment. The random parameters do not need to be tuned and hence reduces the search space.
	\item RandAugStack: RandAug followed by Gaussian blur. 
\end{itemize} 

For the common {\it random resized crop} used along with the above three strategies, work of \cite{ouyang2016factors} explains that the optimal hyper-parameter for random resized crop is (0.2,1) in self-supervised contrastive learning. This setting is also adopted by other work of \cite{DBLP:conf/icml/ChenK0H20, DBLP:conf/cvpr/He0WXG20, DBLP:journals/corr/abs-2003-04297, DBLP:conf/nips/GrillSATRBDPGAP20, DBLP:conf/nips/CaronMMGBJ20}. However, in this paper, we observe severe performance degradation on ImageNet-LT with ResNet-50 (55.0\% vs 52.2\%) for PaCo when we change the hyper-parameter from (0.08,1) to (0.2, 1).
This is because PaCo involves center learning while other self-supervised frameworks only apply non-parametric contrastive loss as described in Section \ref{Sec:contrastive_learning}.
Note that the same phenomenon is also observed on traditional supervised learning with cross-entropy loss.

Another observation is that GPaCo can make better use of strong augmentations compared with PaCo. The work of \cite{wang2021contrastive} demonstrates that directly applying strong data augmentation in MoCo \cite{DBLP:conf/cvpr/He0WXG20, DBLP:journals/corr/abs-2003-04297} does not work well. Here we observe a similar conclusion with RandAug \cite{DBLP:conf/nips/CubukZS020} for PaCo. However, after removing the momentum encoder, GPaCo can achieve better performance with stronger augmentation as shown in Table~\ref{tab:ablation_strongaug}. 

\begin{table}[t]
	\centering
	\setlength{\tabcolsep}{10pt}
	\caption{\textbf{Ablation for the necessary of two-view training on full ImageNet with ViT-B.}}
	\label{tab:ablation_twoview}
	{
		\begin{tabular}{lcc}
			\toprule
			Method  &Full ImageNet($\uparrow$) &ImageNet-C(rel. mCE $\downarrow$) \\
			\midrule
			single-view   &82.5 & 49.1\\
			two-view      &84.0 & 47.3\\
			\bottomrule
		\end{tabular}
	}
\end{table}

\begin{table}[t]
	\centering
	\setlength{\tabcolsep}{10pt}
	\caption{\textbf{Ablation for model ensemble on ImageNet-LT.} ``\dag" represents the ensemble model of a ResNeXt-50 and a ResNeXt-101. ``*" represents the results are from their original paper \cite{9774921}.}
	\label{tab:ablation_ensemble}
	\resizebox{1.0\linewidth}{!}
	{
		\begin{tabular}{lcccc}
			\toprule
			Method  &Many &Medium &Few &All\\
			\midrule
			\multicolumn{5}{c}{\textbf{Single Model}} \\
			\midrule
			ResLT *(ResNeXt-50)    &63.0 &50.5 &35.5 &52.9 \\
			RIDE (ResNeXt-50)      &67.2 &49.0 &28.1 &53.2 \\
			GPaCo (ResNeXt-50)     &67.4 &57.1 &41.2 &58.9 \\
			GPaCo (ResNeXt-101)    &68.7 &59.5 &42.8 &60.8 \\
			\midrule
			\multicolumn{5}{c}{\textbf{Model Ensemble}} \\
			\midrule
			ResLT*           &64.0 &56.6 &44.8 &57.6 \\
			RIDE             &71.8 &53.9 &32.0 &57.8 \\
			GPaCo (ResNeXt-50 2-experts)           &69.9 &59.5 &43.4 &61.3 \\
			GPaCo \dag                             &70.1 &61.2 &45.0 &62.4 \\
			GPaCo (ResNeXt-101 2-experts)          &70.9 &62.0 &45.2 &\textbf{63.2} \\
			\bottomrule
		\end{tabular}
	}
\vspace{-0.1in}
\end{table}

\vspace{+0.1in}
\noindent {\bf Multi-task Learning.}
Re-weighting is a classical method for dealing with imbalanced data. Here we directly apply the re-weighting method of Cui~\cite{cb-focal} in contrastive learning to compare with PaCo. Moreover, Balanced softmax (BalSfx) \cite{DBLP:conf/nips/RenYSMZYL20}, as one state-of-the-art method for traditional cross-entropy in long-tailed recognition, is also applied to contrastive learning rebalance. The experimental results are summarized in Table~\ref{tab:ablation_reweighting_supp}. It is obvious PaCo significantly surpasses the two baselines.

PaCo/GPaCo balances the contrastive learning (for moderating contrast among samples). However the center learning also needs to be balanced, which has been explored in \cite{DBLP:journals/nn/BudaMM18,DBLP:conf/cvpr/HuangLLT16,DBLP:conf/cvpr/CuiJLSB19, he2009learning,chawla2002smote, shen2016relay, DBLP:conf/nips/RenYSMZYL20, DBLP:conf/iclr/KangXRYGFK20, DBLP:journals/corr/abs-2010-01809, DBLP:journals/corr/abs-2101-10633, DBLP:conf/nips/TangHZ20,duggal2020elf}. To compare with state-of-the-art methods in long-tailed recognition, we incorporate Balanced Softmax (BalSfx) \cite{DBLP:conf/nips/RenYSMZYL20} into the center learning. As shown in Table~\ref{tab:ablation_center_rebalance_supp}, after rebalance in center learning, GPaCo boosts performance to 58.9\%, surpassing baselines. 

We also verify the advantage of GPaCo over multi-task learning on balanced data --- full ImageNet and full CIFAR-100. As shown in Table~\ref{tab:ablation_multitask_cifar100}, extensive values for the weight between cross-entropy and supervised contrastive learning have been explored. GPaCo obviously achieve much higher performance.

\begin{figure}[t]
	\begin{center}
		\includegraphics[width=0.8\linewidth]{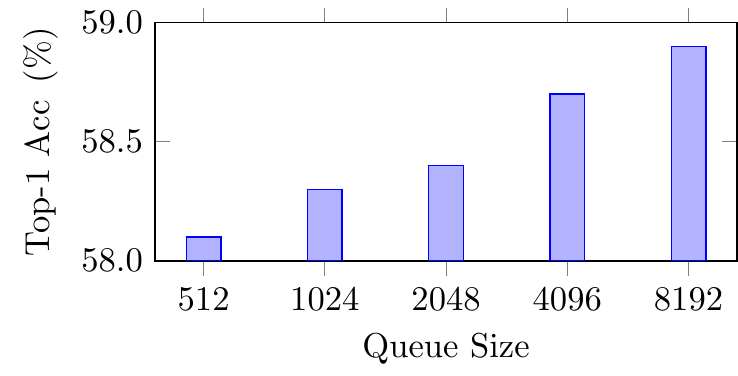}
             \vspace{-0.1in}
		\caption{\textbf{The larger queue size, the better performance.} Experimental results with ResNeXt-50 on ImageNet-LT are plotted. }
		\label{fig:ablation_queuesize}
	\end{center}
\vspace{-0.1in}
\end{figure}

\vspace{+0.1in}
\noindent {\bf GPaCo with Queue Length.}
The queue is designed to enlarge the number of samples for GPaCo/PaCo loss. We examine the effects of different queue size on model performance.
Empirical study on ImageNet-LT with ResNeXt-50 is conducted. As shown in Fig~\ref{fig:ablation_queuesize}, the larger queue size usually leads to better performance. However, when the queue size increase from 4096 to 8192, the performance gains become much smaller.

\vspace{+0.1in}
\noindent {\bf Necessary of Two Views in GPaCo.}
In training of GPaCo, a positive pair consists of two images belonging to the same class. We explore whether it is necessary to generate two views as in the self-supervised learning \cite{DBLP:conf/icml/ChenK0H20, DBLP:conf/cvpr/He0WXG20,DBLP:journals/corr/abs-2003-04297}. As shown in Table~\ref{tab:ablation_twoview}, performance and robustness significantly drops with single view training, which implies the vast importance of two views training in GPaCo.

\vspace{+0.1in}
\noindent{\bf Model Ensemble.}
Models trained on imbalanced data can easily suffer from over-fitting issue on low-frequency classes, thus leading to high variance of model predictions. RIDE proposes an improved version of model ensemble strategy and has verified the advantages of model ensemble for long-tailed recognition. However, GPaCo/PaCo improves single model performance with better feature representation. We go deeper to explore the model ensemble strategy on GPaCo models. As shown in Table \ref{tab:ablation_ensemble},  the improvements coming from model ensemble is much smaller for GPaCo models when compared with RIDE and ResLT (improvements of ResLT v.s. RIDE v.s. GPaCo are 4.7\% v.s. 4.6\% v.s. 2.4\% ), demonstrating that GPaCo training can reduce prediction variance of trained models.

%% file: doc/conclu.tex
\vspace{-0.1in}
\section{Conclusion}
In this paper, we have proposed the Generalized Parametric Contrastive Learning (GPaCo/PaCo), which can deal with both imbalanced and balanced data well. It is based on the theoretical analysis of supervised contrastive learning and rebalance from the convergent optimal values. On balanced data, our analysis of PaCo demonstrates that it can adaptively enhance the intensity of pushing two samples of the same class close as more samples are pulled together with their corresponding centers, which can potentially benefit hard examples learning in training. 

We conduct experiments on various benchmarks of CIFAR-LT, ImageNet-LT, Places-LT, and iNaturalist 2018. The experimental results show that we create a new state-of-the-art for long-tailed recognition. With balanced data, experimental results on full ImageNet and CIFAR show that GPaCo models have better generalization ability and stronger robustness. Transfering GPaCo to semantic segmentation, obvious improvements are obtained.